\documentclass{article}
\usepackage{ijcai20}

\usepackage{times}
\usepackage{soul}
\usepackage{url}
\usepackage[hidelinks]{hyperref}
\usepackage[utf8]{inputenc}
\usepackage[small]{caption}
\usepackage{graphicx}
\usepackage{amsmath}
\usepackage{amsthm}
\usepackage{booktabs}
\usepackage{algorithm}
\usepackage{algorithmic}
\usepackage{amssymb}

\title{Neural Belief Reasoner\thanks{Published as a conference paper at the International Joint Conference on Artificial Intelligence (IJCAI) 2020.
    Official link at \url{https://doi.org/10.24963/ijcai.2020/590}.
    The sole copyright holder is IJCAI, all rights reserved.}}
\renewcommand\footnotemark{}
\author{
    Haifeng Qian
    \affiliations
    IBM Research, Yorktown Heights, NY, USA
    \emails
    qianhaifeng@us.ibm.com
}

\begin{document}

\maketitle

\begin{abstract}
This paper proposes a new generative model called neural belief reasoner (NBR).
It differs from previous models in that it specifies a belief function rather than a probability distribution.
Its implementation consists of neural networks, fuzzy-set operations and belief-function operations, and query-answering, sample-generation and training algorithms are presented.
This paper studies NBR in two tasks.
The first is a synthetic unsupervised-learning task, which demonstrates NBR's ability to perform multi-hop reasoning, reasoning with uncertainty and reasoning about conflicting information.
The second is supervised learning: a robust MNIST classifier for 4 and 9, which is the most challenging pair of digits.
This classifier needs no adversarial training, and it substantially exceeds the state of the art in adversarial robustness as measured by the $L_2$ metric, while at the same time maintains 99.1\% accuracy on natural images.
\end{abstract}

\section{Introduction} \label{sec:intro}

It is a widely held hypothesis that bridging the gap between machine learning and rule-based reasoning would bring benefits to both domains.
For rule-based systems, this could provide an elegant solution to automatically discover rules from observations, and could provide new approaches to reasoning with uncertainty.
On the other hand, despite the phenomenal successes of deep learning, neural networks tend to have poor robustness and interpretability, both of which are nonissue in rule-based systems.
The robustness issue has recently been highlighted by the existence of adversarial examples in many systems.
For example, even for the well-studied MNIST task, the state of the art in $L_2$ robustness is far from satisfactory and comes at a cost to accuracy on natural images.

Some of the works at the intersection of machine learning and reasoning will be reviewed in Section~\ref{sec:liter}.
A prominent one is Boltzmann machine and its variants \cite{dbm}, which combine neural networks and Markov random fields.
A recent example is differentiable inductive logic programming \cite{dilp}, which combines neural networks and logic programs.

This paper presents a new approach called neural belief reasoner (NBR), which combines neural networks and belief functions.
Belief functions are a generalization of probability functions \cite{ds}, and have the advantage of modeling epistemic uncertainty, i.e., the lack of knowledge, and an elegant way of combining multiple sources of information.
Despite these advantages, belief functions have seen much less adoption than mainstream methods like Bayesian networks and Markov random fields.
NBR is built on two innovations: 1) using neural networks to represent fuzzy sets, and 2) using fuzzy sets to specify a belief function.
From the machine-learning perspective, NBR is a new generative model that specifies a belief function rather than a probability distribution.
From the rule-based perspective, NBR is a new method of reasoning with uncertainty that enables automatic discovery of non-symbolic rules from observations and that uses belief functions to model uncertainty.

The next section will define the model and present query-answering, sample-generation and training algorithms.
Then Sections~\ref{sec:toy} and \ref{sec:mnist} demonstrate NBR's capabilities through two tasks.
The first task is unsupervised learning: in a synthetic 11-bit world where only partial observations are available, an NBR model is trained and then answers queries.
The queries involve multi-hop reasoning, reasoning with uncertainty and reasoning about conflicting information.
The second task is supervised learning: a robust MNIST classifier for 4 and 9, which is the most challenging pair of digits.
It sets a new state of the art in adversarial robustness as measured by the $L_2$ metric and maintains 99.1\% accuracy on natural images.

\section{Neural Belief Reasoner} \label{sec:tech}

Let's start with a restricted framework of reasoning in Section~\ref{sec:proto}, which serves as a stepping stone towards NBR's definitions in Section~\ref{sec:def} and algorithms after that.

\subsection{Prototype with Classical Sets} \label{sec:proto}

Let $U$ denote the sample space, i.e., the set of all possibilities.
Consider a reasoning framework where a model is composed of $K$ sets, $R_1,\cdots,R_K \subseteq U$, and each $R_i$ is annotated with a scalar $0\leq b_i \leq 1$.
Let's interpret each $R_i$ as a logic rule: an outcome $x\in U$ is said to satisfy $R_i$ if and only if $x \in R_i$.
Let's interpret $b_i$ as the belief in $R_i$.

Define vector ${\bf y} \triangleq \left( y_1,\cdots,y_K \right)$ where entries are 0 or 1.
Define intersection sets $S_{\bf y} \triangleq \bigcap_{1\leq i \leq K \mid y_i=1} R_i$, and define $S_{\left( 0,\cdots,0 \right)} \triangleq U$.
Intuitively each $S_{\bf y}$ contains outcomes that satisfy a subset of the $K$ rules as selected by ${\bf y}$.
Define scalar function $p\left({\bf y}\right) \triangleq \prod_{i=1}^K\left(b_i \cdot y_i+\left(1-b_i\right)\cdot\left(1-y_i\right)\right)$.

Let $2^U$ denote the power set of $U$. Define the following function from $2^U$ to $\mathbb{R}$: $m\left(\emptyset\right) \triangleq 0$; for $A\neq\emptyset$,
\begin{equation} \label{eq:mproto}
m\left(A\right) \triangleq  \frac{\sum_{{\bf y}\mid S_{\bf y}=A}p\left({\bf y}\right)}{1-\sum_{{\bf y}\mid S_{\bf y}=\emptyset}p\left({\bf y}\right)}
\end{equation}
It is straightforward to verify that $\sum_{A\subseteq U}m\left(A\right)=1$.
Therefore this function $m\left(\cdot\right)$ satisfies the requirements to be a basic probability assignment \cite{ds}.
Hence this function $m\left(\cdot\right)$ uniquely specifies a belief function over $U$ \cite{ds}: $\mathrm{Bel}\left(A\right) = \sum_{B\subseteq A}m\left(B\right),\forall A\subseteq U$.

Intuitively this framework considers $2^K$ possible worlds: each world corresponds to each ${\bf y}$, and in each world a subset of the $K$ rules, as selected by ${\bf y}$, exist.
Each satisfiable world, i.e., where $S_{\bf y} \neq \emptyset$, is assigned a mass that is proportional to $p\left({\bf y}\right)$, the product of $b_i$'s for rules that are present and $\left(1-b_i\right)$'s for rules that are absent.
Each unsatisfiable world, i.e., where $S_{\bf y} = \emptyset$, is assigned a mass of zero.
The total mass is one, which is achieved through the denominator in (\ref{eq:mproto}) that is essentially Dempster's rule of combination \cite{ds}.

With a belief function defined, this framework is able to answer queries.
Similar to conditional probabilities in traditional models, it answers with conditional belief functions.
Given a condition $C\subseteq U$ and a proposition $Q\subseteq U$, the conditional belief and conditional plausibility are
\begin{equation} \label{eq:qproto}
\begin{aligned}
  \mathrm{Bel}\left(Q\mid C \right) & = \frac{\mathrm{Bel}\left(Q\cup\overline{C}\right)-\mathrm{Bel}\left(\overline{C}\right)}{1-\mathrm{Bel}\left(\overline{C}\right)} \\
  & = 1- \frac{\sum_{{\bf y}\mid S_{\bf y}\cap C\cap\overline{Q}\neq \emptyset}p\left({\bf y}\right)}{\sum_{{\bf y}\mid S_{\bf y}\cap C\neq \emptyset}p\left({\bf y}\right)} \\
  \mathrm{Pl}\left(Q\mid C \right) & = 1-\mathrm{Bel}\left(\overline{Q}\mid C \right) \\
  & = \frac{\sum_{{\bf y}\mid S_{\bf y}\cap C\cap Q\neq \emptyset}p\left({\bf y}\right)}{\sum_{{\bf y}\mid S_{\bf y}\cap C\neq \emptyset}p\left({\bf y}\right)}
\end{aligned}
\end{equation}
Intuitively, $\mathrm{Bel}\left(\cdot\right)$ quantifies the evidence that supports a proposition and $1-\mathrm{Pl}\left(\cdot\right)$ quantifies evidence against it.

\subsection{Model Definitions} \label{sec:def}

I now define the full form of NBR by generalizing the previous section in a number of ways: $R$'s are replaced by fuzzy sets represented by neural networks; $U$ becomes the latent space which is separated from observation space.

\begin{figure}[t]
\centering
\includegraphics[width=0.99\columnwidth]{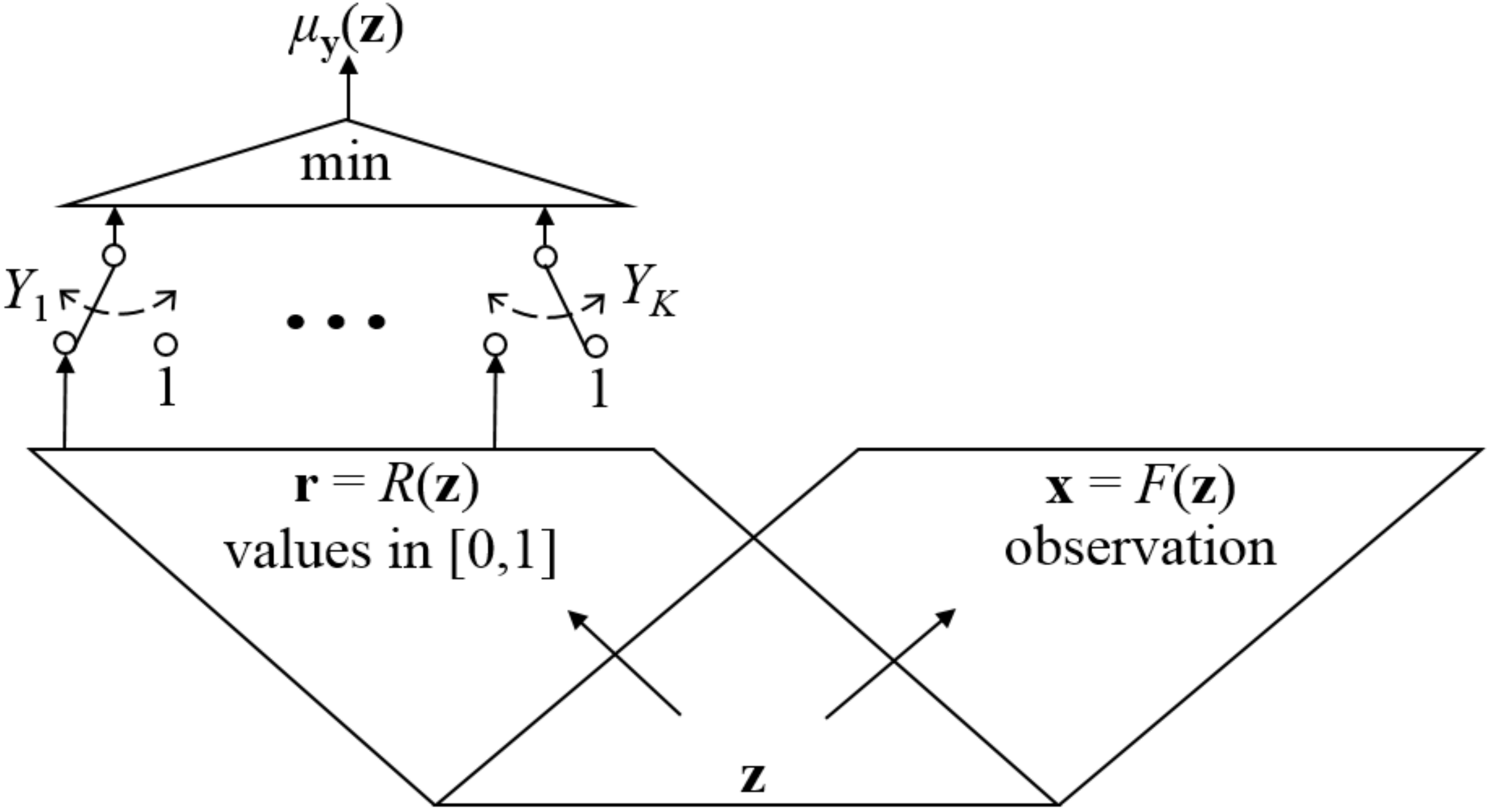}
\caption{Architecture of NBR.}
\label{fig:nbr}
\end{figure}

An NBR model has the following components and Figure~\ref{fig:nbr} illustrates the architecture.
\begin{itemize}
\item
Function ${\bf x}=F\left({\bf z}\right)$ where vector $\bf x$ is the observation variables and vector {\bf z} is the latent variables.
\item
Function ${\bf r}=R\left({\bf z}\right)$ with output values in range $\left[0,1\right]$.
\item
Bernoulli variables ${\bf Y}=\left(Y_1,Y_2,\cdots,Y_K\right)$, where $K$ is the dimension of $\bf r$.
\end{itemize}
The $F$ and $R$ functions can be implemented by neural networks.
The parameters of an NBR model are the parameters of functions $F$ and $R$, and $b_i = P_{Y_i}(1)$ for $i=1,2,\cdots,K$.

Each $R_i\left({\bf z}\right)$ is interpreted as the membership function of a fuzzy set over ${\bf z}$ space \cite{fuzzy}.
I consider the same $2^K$ possible worlds as in Section~\ref{sec:proto}, but the intersection set $S_{\bf y}$ for each world now becomes the intersection of fuzzy sets and has the following membership function:
\begin{equation} \label{eq:member}
\mu_{\bf y}\left({\bf z}\right) = \min_{i=1}^K\left(y_i\cdot R_i\left({\bf z}\right)+1-y_i\right)
\end{equation}
Consequently, the satisfiability of each world is no longer a binary property, but a degree in $\left[0,1\right]$: $\max_{\bf z}\mu_{\bf y}\left({\bf z}\right)$.
Therefore, the mass assigned to each world should be proportional to $p\left({\bf y}\right)\cdot \max_{\bf z}\mu_{\bf y}\left({\bf z}\right)$.
I still need to ensure that the total mass is one, and that leads to the following formula which replaces (\ref{eq:mproto}) as the new basic probability assignment:
\begin{equation} \label{eq:m}
m\left(S_{\bf y}^\prime\right) \triangleq  \frac{p\left({\bf y}\right)\cdot \max_{\bf z}\mu_{\bf y}\left({\bf z}\right)}{\sum_{\bf y^\prime}\left( p\left({\bf y^\prime}\right)\cdot \max_{\bf z}\mu_{\bf y^\prime}\left({\bf z}\right)\right)}
\end{equation}
where $S_{\bf y}^\prime$ is a fuzzy set with the membership function of $\mu_{\bf y}\left({\bf z}\right)/\max_{\bf z^\prime}\mu_{\bf y}\left({\bf z^\prime}\right)$: $S_{\bf y}^\prime$ is scaled $S_{\bf y}$ such that at least one point in the ${\bf z}$ space is fully included.
Considering that $p\left(\cdot\right)$ is exactly the probability function of the Bernoulli vector ${\bf Y}$, the above has a more concise form:
\begin{equation} \label{eq:m2}
m\left(S_{\bf y}^\prime\right) \triangleq \frac{p\left({\bf y}\right)\cdot \max_{\bf z}\mu_{\bf y}\left({\bf z}\right)}{\mathrm{E}\left[\max_{\bf z}\mu_{\bf Y}\left({\bf z}\right)\right]}
\end{equation}
With this new $m\left(\cdot\right)$ function, a belief function is specified over the ${\bf z}$ space: for any fuzzy set $A$,
\begin{equation}\label{eq:fuzzybel}
\mathrm{Bel}\left(A\right) \triangleq \sum_{\bf y}m\left(S_{\bf y}^\prime\right)\cdot\left(1-\max_{\bf z}\mu_{S_{\bf y}^\prime\cap\overline{A}}\left({\bf z}\right)\right)
\end{equation}
where $\mu_{S_{\bf y}^\prime\cap\overline{A}}\left({\bf z}\right) \triangleq \min\left(\mu_{\bf y}\left({\bf z}\right)/\max_{\bf z^\prime}\mu_{\bf y}\left({\bf z^\prime}\right),1-\mu_A\left({\bf z}\right)\right)$ and where $\mu_A\left({\bf z}\right)$ is the membership function of $A$.

\subsection{Query Answering} \label{sec:query}

The general form of a query is a conditional belief function given a condition function $C\left({\bf z}\right)$ which outputs a scalar in range $\left[0,1\right]$.
The condition may come as a function of ${\bf x}$, and since ${\bf x}$ is a deterministic function of ${\bf z}$, $C\left({\bf z}\right)$ is the general form and is the membership function of a fuzzy set in ${\bf z}$ space.

To answer a query, I add $C\left({\bf z}\right)$ as an extra entry to ${\bf r}$, and add an extra entry of constant 1 to ${\bf Y}$.
Intuitively, the NBR has an additional rule that always exists.
After such additions, formulas (\ref{eq:m2})(\ref{eq:fuzzybel}) specify a conditional belief function.

Let us consider a special type of queries that are the most common in practice: given a Boolean function $C\left({\bf z}\right)$ as condition, compute the conditional belief and plausibility of another Boolean function $Q\left({\bf z}\right)$.
In other words, I look for the replacement of (\ref{eq:qproto}).
The following formulas can be derived from (\ref{eq:m2})(\ref{eq:fuzzybel}) and the proof is omitted due to space limit.
\begin{align}
\mathrm{Bel}\left( Q\left(\cdot\right) \mid C\left(\cdot\right) \right) & = 1 - 
\frac{\mathrm{E}\left[ \displaystyle\max_{{\bf z} \mid C\left({\bf z}\right)=1,Q\left({\bf z}\right)=0} \mu_{\bf Y}\left({\bf z}\right)\right]}
{\mathrm{E}\left[ \displaystyle\max_{{\bf z} \mid C\left({\bf z}\right)=1} \mu_{\bf Y}\left({\bf z}\right)\right]} \label{eq:bel} \\
\mathrm{Pl}\left( Q\left(\cdot\right) \mid C\left(\cdot\right) \right) & =
\frac{\mathrm{E}\left[ \displaystyle\max_{{\bf z} \mid C\left({\bf z}\right)=1,Q\left({\bf z}\right)=1} \mu_{\bf Y}\left({\bf z}\right)\right]}
{\mathrm{E}\left[ \displaystyle\max_{{\bf z} \mid C\left({\bf z}\right)=1} \mu_{\bf Y}\left({\bf z}\right)\right]} \label{eq:pl}
\end{align}

\subsection{Sample Generation} \label{sec:sample}

A belief function allows sample generation only if it is also a probability function.
When combining two belief functions where one of the two is a probability function, by Dempster's rule of combination \cite{ds}, the resulting belief function is always a probability function.
Therefore, to generate observation samples like traditional generative models, I combine the belief function of an NBR with a probability function over the ${\bf x}$ space.
This probability function is referred to as the \emph{prior-knowledge distribution}.
Intuitively, the prior-knowledge distribution represents assumptions or knowledge that are not included in this NBR.
For example, if one only knows the range of ${\bf x}$, a uniform prior-knowledge distribution can be used; if one knows the mean and variance of ${\bf x}$, a Gaussian distribution can be used.
As will become evident later, I only require the ability to draw samples from it.
The flexibility to combine an NBR with various prior-knowledge distributions is analogous to applying the same knowledge in multiple environments.

For clarity of presentation, let us focus on the scenario where the ${\bf x}$ space is discrete.
For a sample value ${\bf \tilde{x}}$, let $P_0\left({\bf \tilde{x}}\right)$ denote its probability in the prior-knowledge distributions.
Its probability after combining with an NBR is
\begin{equation} \label{eq:gen}
P\left( {\bf x} = {\bf \tilde{x}} \right) = \frac{P_0\left({\bf \tilde{x}}\right) \cdot \mathrm{Pl}\left( {\bf x} = {\bf \tilde{x}} \right)}
{\sum_{\bf x^\prime}\left( P_0\left({\bf x^\prime}\right) \cdot \mathrm{Pl}\left( {\bf x} = {\bf x^\prime} \right) \right)}
\end{equation}
where the plausibilities are given by (\ref{eq:pl}) with $C$ being always true.
To generate samples according to (\ref{eq:gen}), I draw samples from the prior-knowledge distribution and randomly keep or discard a sample, such that the probability to keep sample ${\bf \tilde{x}}$ is proportional to $\mathrm{Pl}\left( {\bf x} = {\bf \tilde{x}} \right)$.
Note that the denominator in (\ref{eq:pl}) does not change with $Q$ and hence is the same for all ${\bf \tilde{x}}$ values; therefore I can simply use the numerator.
In summary, the probability to keep a sample ${\bf \tilde{x}}$ is:
\begin{equation} \label{eq:sample}
P_\mathrm{keep}\left( {\bf \tilde{x}} \right) =
\mathrm{E}\left[ \displaystyle\max_{{\bf z} \mid F\left({\bf z}\right)={\bf \tilde{x}}} \mu_{\bf Y}\left({\bf z}\right) \right].
\end{equation}

When the ${\bf x}$ space is continuous, the generation procedure is the same: draw samples from a continuous prior-knowledge distribution and randomly keep or discard a sample according to the keep probability of (\ref{eq:sample}).

\subsection{Training} \label{sec:train}

For unsupervised learning, an NBR is trained by maximizing the likelihood of observations in the sample generation process of the previous section.
Therefore one needs to choose a prior-knowledge distribution for training.
This choice defines what should be learned by the NBR: new knowledge that is present in the observations yet that is not already encoded in the prior-knowledge distribution.
For example, samples generated by an existing NBR can be used as the prior-knowledge distribution to train a new NBR, and then this new NBR would be trained to learn and only learn new knowledge that is not in that existing NBR.
Note that, after an NBR is trained, the query-answering process of Section~\ref{sec:query} is independent of the prior-knowledge distribution used in training.
In other words, an NBR's answers are based on only the knowledge contained in itself, and this enables modular and transferable knowledge representation.

Given observations ${\bf x_1},\cdots,{\bf x_n}$, the likelihood loss is
\begin{equation} \label{eq:likeli}
\begin{aligned}
\mathcal{L} = & -\frac{1}{n} \sum_{i=1}^{n} \log P\left( {\bf x} = {\bf x_i} \right) \\
  = &  -\frac{1}{n} \sum_{i=1}^{n} \log \frac{ P_0\left({\bf x_i}\right) \cdot P_\mathrm{keep}\left( {\bf x_i} \right) }
{\sum_{\bf x^\prime}\left( P_0\left({\bf x^\prime}\right) \cdot P_\mathrm{keep}\left( {\bf x^\prime} \right) \right)} \\
  = &  -\frac{1}{n} \sum_{i=1}^{n} \log P_0\left({\bf x_i}\right) -\frac{1}{n} \sum_{i=1}^{n} \log P_\mathrm{keep}\left( {\bf x_i} \right) \\
    & + \log\sum_{\bf x^\prime}\left( P_0\left({\bf x^\prime}\right) \cdot P_\mathrm{keep}\left( {\bf x^\prime} \right) \right)
\end{aligned}
\end{equation}
The first term is a constant and can be removed; the last term can be shortened by letting ${\bf X}$ denote a random vector that has the prior-knowledge distribution.
The loss function becomes:
\begin{equation} \label{eq:likeli2}
\mathcal{L} = - \frac{1}{n} \sum_{i=1}^{n} \log P_\mathrm{keep}\left( {\bf x_i} \right)
+ \log \mathrm{E}\left[ P_\mathrm{keep}\left( {\bf X} \right) \right]
\end{equation}
Each training iteration uses a batch of observations to approximate the first term and a batch of samples from the prior-knowledge distribution to approximate the second term.
One implementation issue is that the expectation is before log in the second term and small batch size causes bias in gradient estimation.
In practice, I use the following loss instead:
\begin{equation} \label{eq:likeli3}
\mathcal{L} = - \frac{1}{n} \sum_{i=1}^{n} \log P_\mathrm{keep}\left( {\bf x_i} \right)
+ \mathrm{E}\left[ P_\mathrm{keep}\left( {\bf X} \right) \right]/\alpha
\end{equation}
where $\alpha$ is a constant that gets updated once every certain number of batches, and its value is an estimate of $\mathrm{E}\left[ P_\mathrm{keep}\left( {\bf X} \right) \right]$ on a large number of samples.
It is straightforward to verify that, with a large batch size, (\ref{eq:likeli2}) and (\ref{eq:likeli3}) result in asymptotically the same gradients with respect to model parameters.
(\ref{eq:likeli3}) is linear with respect to the expectation in the second term and hence small batch size can be used without causing bias in gradient estimation.

\section{Unsupervised Learning: a Synthetic Task} \label{sec:toy}

This section gives a first demonstration of NBR on an unsupervised-learning task.
Source code for training and inference is available at\\
\url{http://researcher.watson.ibm.com/group/10228}

Consider a world with 11 bits.
Partial observations are available that observe either the first 10 bits or the last 10 bits.
In observations of the first 10 bits, the first bit is the majority function of the middle 9 bits for 90\% of the cases, and the inverse for 10\% of the cases.
In observations of the last 10 bits, the last bit is the majority function of the middle 9 bits for 20\% of the cases, and the inverse for 80\%.

An NBR is trained on these observations: $K$ is 2; $F\left(\cdot\right)$ is identity function; $R\left(\cdot\right)$ is two three-layer ReLU networks, where each network is followed by a sigmoid unit, and where one network takes the first 10 bits as input and the other takes the last 10 bits.
The loss (\ref{eq:likeli3}) is used, and the prior-knowledge distribution is uniform over the $2^{11}$ possibilities.
For a partial observation ${\bf x_i}$, let ${\bf x_{i,0}}$ and ${\bf x_{i,1}}$ be the two possible full observations.
Substituting $P\left( {\bf x} = {\bf x_i} \right)=P\left( {\bf x} = {\bf x_{i,0}} \right)+P\left( {\bf x} = {\bf x_{i,1}} \right)$ into (\ref{eq:likeli}) and following the same derivation to (\ref{eq:likeli3}), it is straightforward to see that I simply need to compute $P_\mathrm{keep}\left( {\bf x_i} \right)$ in (\ref{eq:likeli3}) as $P_\mathrm{keep}\left({\bf x_{i,0}}\right)+P_\mathrm{keep}\left({\bf x_{i,1}}\right)$.

\begin{table}[t]
\centering
\resizebox{.99\columnwidth}{!}{
\begin{tabular}{llcc}
\toprule
$C$ & $Q$ & belief & plausibility \\
\midrule
$x_0=1$ & $x_{10}$ & 0 & 0.33 \\
$x_0=0$ & $x_{10}$ & 0.67 & 1 \\
$x_0=1,x_{1-4}=0$ & $x_5$ & 0.89 & 1 \\
$x_0=1,x_{1-4}=0,x_{10}=1$ & $x_5$ & 0.67 & 1 \\
$x_0=1,x_{1-4}=0,x_{6-10}=1$ & $x_5$ & 0.67 & 0.75 \\
\bottomrule
\end{tabular}
}
\caption{Answers to example queries by NBR.}
\label{tbl:proto}
\end{table}

Table~\ref{tbl:proto} lists NBR's answers to five queries.
The belief values are computed by (\ref{eq:bel}) and the plausibility values are by (\ref{eq:pl}).
In the first query, the condition is the first bit being 1 and the query is on the last bit.
NBR answers with belief zero and plausibility 0.33, which means that it has some evidence to support $x_{10}$ being 0 but no evidence to support $x_{10}$ being 1.
This answer is intuitive: with $x_0=1$, it is likely that the majority of the middle 9 bits is 1, and consequently it is likely that $x_{10}$ is 0.
Recall that during training the NBR has never seen an observation that simultaneously shows $x_0$ and $x_{10}$, and it answers the query by performing multi-hop reasoning with uncertainty.
The second query is the opposite: with $x_0=0$, NBR has some evidence to support $x_{10}$ being 1 but no evidence to support $x_{10}$ being 0.
There is an interesting comparison between the third and fourth queries: NBR's belief decreases when $x_{10}=1$ is added to the condition.
The reason is that $x_0=1$ and $x_{10}=1$ are two conflicting pieces of information, and as a result the denominator in (\ref{eq:bel}) is reduced to less than 1 for the fourth query, which in turn causes the belief value to decrease.
This is consistent with human intuition when facing conflicting information.
The fifth query is a similar case of conflicting information where all bits but $x_5$ are fixed.
NBR's answer means that it has evidence to support both possibilities for $x_5$ and that the evidence for $x_5$ being 1 is stronger than the evidence for 0.
This is again consistent with human intuition based on observations of this world.
It's worth noting that the gap between belief and plausibility narrows for the fifth query, which reflects more information about the world from the condition, however the gap still exists, which reflects epistemic uncertainty, -- the fact that NBR does not have complete knowledge about the world.
The gap only closes when an NBR has complete knowledge regarding a query, in which case the answer is a probability function.

To demonstrate the separation between latent space and observation space, a second NBR model is trained with nontrivial $F\left(\cdot\right)$.
First an autoencoder is trained with a latent space of dimension 8.
I then use the decoder part as $F\left(\cdot\right)$ and fix it as non-trainable during NBR training.
When I need to evaluate the max operation in (\ref{eq:sample}), I feed ${\bf \tilde{x}}$ into the encoder part to compute ${\bf z}$ as a cheap surrogate operation.
The resulting NBR gives the same answers as in Table~\ref{tbl:proto}.

\section{Supervised Learning: a Robust Classifier} \label{sec:mnist}

This section demonstrates NBR for supervised learning, and specifically discusses a robust MNIST classifier for 4 and 9.

\subsection{Using NBR for Classification} \label{sec:classi}

It is possible to convert a classification task to unsupervised learning by treating labels as part of the observation.
However, that is not the most efficient way to use NBR for classification, and this section presents a better approach.
Most discussions are applicable to classification in general.

Given an MNIST image, consider a world with 10 possibilities, -- one of the ten labels is true.
Recall from Section~\ref{sec:def} that the role of each entry in ${\bf r}$ is to specify a fuzzy set.
In a world with 10 possibilities, a fuzzy set is defined by 10 grades of membership, i.e., 10 numbers between 0 and 1.
Therefore, each entry in ${\bf r}$ can be implemented by an arbitrary MNIST classifier, and I simply add 10 sigmoid units at the end to convert logits to values in $\left[0,1\right]$.
Function $F\left(\cdot\right)$ is not used.
Therefore, an NBR classifier is composed of $K$ classifiers with sigmoids added and Bernoulli variables ${\bf Y}$.

Now let's define its outputs.
With (\ref{eq:bel})(\ref{eq:pl}), the belief and plausibility of each label can be computed; let them be $\mathrm{Bel}_j$, $\mathrm{Pl}_j$, $0\leq j\leq 9$.
The output vector in the implementation is:
\begin{equation} \label{eq:classi}
{\bf o} = \left( \log\mathrm{Pl}_0 ,\cdots, \log\mathrm{Pl}_9 \right)
\end{equation}
and its argmax is NBR's output label.
The values in (\ref{eq:classi}) are the negative of weights of evidence against each label \cite{ds}.
There are other choices: for example, $\log\mathrm{Pl}_j-\log\left(1-\mathrm{Bel}_j\right)$ is another reasonable choice which combines weights of evidence for and against label $j$.

\subsection{Distinct Bodies of Evidence} \label{sec:fod}

The intuition behind robust classification is to divide a classification task into simpler tasks, each of which is so simple that it can be solved robustly by a single $R_i\left(\cdot\right)$, and then NBR combines the $K$ sources to solve the overall task.
Dempster's rule requires that sources represent entirely distinct bodies of evidence \cite{ds}.
I achieve this by using different frames of discernment and dividing the training data.

To explain by example, consider a rule with this frame of discernment: \{0\} \{5,6\} \{7\}.
It is built as a classifier with 3 classes and specifies a fuzzy set with these grades of membership: $v_1,1,1,1,1,v_2,v_2,v_3,1,1$.
Note that the grades for \{1,2,3,4,8,9\} are always 1; intuitively this rule makes no judgment about labels outside its frame of discernment.
Also note that the grades for 5 and 6 are always the same; intuitively this rule does not distinguish between them.
With different frames of discernment, an NBR can gather enough knowledge for the overall task.

However, some minimal frame of discernment, for example \{4\} \{9\}, is still too complex to solve robustly with a single neural network.
I will focus on this pair and will build an NBR with 2632 rules to classify 4 and 9.
The first 14 rules are neural networks that are trained on different subsets of the training data, while the rest are memorization rules.

Let $T_4$ be the set of training images with label 4, and let $T_9$ be that for 9.
Let $T_{4,i}$, $1\leq i\leq 8$ be mutually exclusive and collectively exhaustive subsets of $T_4$, and let $T_{9,i}$, $1\leq i\leq 8$ be such subsets of $T_9$.
For $1\leq i\leq 7$, the $i^\mathrm{th}$ rule is a binary classifier that is trained to distinguish $T_{4,i}$ versus $T_9$.
For $8\leq i\leq 14$, the $i^\mathrm{th}$ rule is a binary classifier that is trained to distinguish $T_{9,i-7}$ versus $T_4$.

All rules have the form of $\mathrm{sigmoid}\left(s_i\cdot G_i\left(\mathrm{image}\right)\right),1\leq i\leq 2632$, where $s_i$ is a trainable scalar and $G_i\left(\cdot\right)$ is a trainable function that outputs a scalar.
For the first 14 rules, $G_i\left(\cdot\right)$ is a $L_2$-nonexpansive neural network (L2NNN) \cite{l2nnn}.
Each $T_{4,1\leq i\leq 7}$ is $\left\{{\bf t}\in T_4 | G_i\left({\bf t}\right)\geq\gamma\,\mathrm{and}\, G_i\left({\bf t}\right)\geq G_j\left({\bf t}\right),\forall 1\leq j\leq 7\right\}$, where $\gamma$ is a hyperparameter.
Intuitively, each digit 4 is assigned to the $G_i\left(\cdot\right)$ that classifies it the most robustly, and it is assigned to $T_{4,8}$ if none of the seven $G_i\left(\cdot\right)$'s reaches threshold $\gamma$.
The $T_{9,i}$ subsets are similarly defined.
All subsets are periodically updated during the training process.

After the first 14 rules are trained, $T_{4,8}$ contains 653 images and $T_{9,8}$ contains 1965.
These images are handled by memorization rules\footnote{Effects of the memorization rules: if they are removed, the nominal accuracy of the NBR classifier drops from 99.1\% to 98.0\%, while its robust accuracy increases from 55.3\% to 59.1\%.}: 653 rules to distinguish each image in $T_{4,8}$ versus $T_9$, and another 1965 rules for $T_{9,8}$ against $T_4$.
For these rules, $G_i\left({\bf x}\right)\triangleq d_i- \| {\bf x}-{\bf t_i} \|_2,15\leq i\leq 2632$, where ${\bf t_i}$ is the image to memorize and $d_i$ is a trainable scalar; note that this function is nonexpansive with respect to $L_2$.

A final adjustment is needed to produce $R_i\left(\cdot\right)$.
Let $T_i^\prime$ and $T_i^{\prime\prime}$ be the two sets of training images that the $i^\mathrm{th}$ rule is trained to distinguish.
For example, $T_i^\prime$ may be one of the $T_{4,i}$'s while $T_i^{\prime\prime}$ may be $T_9$.
If the sigmoid outputs 0 for image ${\bf t}$, the knowledge obtained is that ${\bf t}$ is dissimilar to those specific 4's in $T_i^\prime$, not all digits 4; hence this rule should not put the plausibility of label 4 to zero.
Therefore I have $R_i\left({\bf t}\right) \triangleq \mathrm{sigmoid}\left(s_i\cdot G_i\left({\bf t}\right)\right)$ if $G_i\left({\bf t}\right) \geq 0$, and otherwise $R_i\left({\bf t}\right) \triangleq 0.5-\frac{\left|T_i^\prime \right|}{\left|T_4 \right|}\cdot\left(0.5-\mathrm{sigmoid}\left(s_i\cdot G_i\left({\bf t}\right)\right)\right)$.

\subsection{Scaling Trick for Linear Complexity}

Let us consider a single rule and let ${\bf v}\triangleq\left(v_1,\cdots,v_J\right)$ be the grades of membership that this rule computes for a particular image over its frame of discernment.
Let $v_\mathrm{max}$ and $v_\mathrm{min}$ denote the max and min grades among them.
Let $b$ denote the corresponding $b_i$ parameter.
If I pretend that this is a single-rule NBR and apply (\ref{eq:m2}), I effectively replace ${\bf v}$ with ${\bf v^\prime}\triangleq\left(v_1/v_\mathrm{max},\cdots,v_J/v_\mathrm{max}\right)$ and replace $b$ with $b^\prime\triangleq b\cdot v_\mathrm{max}/\left(1-b+b\cdot v_\mathrm{max}\right)$.
Note that the max grade in ${\bf v^\prime}$ is always 1.
These replacements do not modify the single-rule NBR at all: (\ref{eq:fuzzybel}) for any set $A$ computes the same value before and after the replacements.

Let us push one step further and also scale the min grade to zero.
Specifically, I replace ${\bf v}$ with ${\bf v^{\prime\prime}}\triangleq\left(\frac{v_1-v_\mathrm{min}}{v_\mathrm{max}-v_\mathrm{min}},\cdots,\frac{v_J-v_\mathrm{min}}{v_\mathrm{max}-v_\mathrm{min}}\right)$ and replace $b$ with $b^{\prime\prime}\triangleq b\cdot \left(v_\mathrm{max}-v_\mathrm{min}\right)/\left(1-b+b\cdot v_\mathrm{max}\right)$.
These replacements subtly modify the single-rule NBR: (\ref{eq:fuzzybel}) is not affected for any classical set $A$, however there is no guarantee if $A$ is a fuzzy set.
It is arguable that such changes are acceptable.

The benefit of using ${\bf v^{\prime\prime}}$ and $b^{\prime\prime}$ is that, if $J=2$, i.e., if the frame of discernment is binary, this rule becomes a classical rule.
If all rules in an NBR are classical, (\ref{eq:bel})(\ref{eq:pl}) become the same as (\ref{eq:qproto}) and can be evaluated by the original Dempster's rule.
Since Dempster's rule is associative \cite{ds}, the worst-case complexity is $O\left(K\cdot 2^L\right)$ where $K$ is the number of rules and $L$ is the number of classes.
Consequently NBR can use a large $K$.
If $L$ is large, a classification task can be done hierarchically.
Note that using only binary frames of discernments is not a strong restriction: for example, \{1,4,7,9\} \{2,3,5,8\} is a binary frame and is expressive.

\subsection{Loss Functions for Robustness} \label{sec:loss}

The training process has three steps.
In the first step, $G_i\left(\cdot\right)$ functions in the first 14 rules are learned.
The first seven $G_i\left(\cdot\right)$'s are trained jointly with the following loss function:
\begin{equation} \label{eq:loss1}
\resizebox{.89\linewidth}{!}{$
\begin{aligned}
\mathcal{L}_{1-7} = & - \sum_{i=1}^7\sum_{{\bf t}\in T_{4,i}}\log\left(\mathrm{sigmoid}\left(s\cdot \left(G_i\left({\bf t}\right)-\beta\right)\right)\right)\\
& - \sum_{{\bf t}\in T_9}\log\left(1-\mathrm{sigmoid}\left(s\cdot \left(\max_{i=1}^7G_i\left({\bf t}\right)+\beta\right)\right)\right)
\end{aligned}
$}
\end{equation}
where $s$ and $\beta$ are hyperparameters.
With the definition of $T_{4,1\leq i\leq 7}$ from Section~\ref{sec:fod}, (\ref{eq:loss1}) can be viewed as the cross-entropy loss of the classifier of $\mathrm{sigmoid}\left(s\cdot \max_{i=1}^7G_i\left({\bf t}\right)\right)$ with a twist: I reduce $G_i\left(\cdot\right)$'s by $\beta$ for digits 4 and increase them by $\beta$ for digits 9.
Intuitively, because $G_i\left(\cdot\right)$'s are L2NNNs, these adjustments realize the worst-case scenario of an adversarial attack of $L_2$ distortion of $\beta$.
This technique is computationally much cheaper than adversarial training, and I refer to it as \emph{poor man's adversarial training}.
Functions $G_i\left(\cdot\right)$ for $8\leq i\leq 14$ are trained with a similar loss.

In the second step, $s_i,1\leq i\leq 2632$ and $d_i,15\leq i\leq 2632$ are learned.
The loss function for the $i^\mathrm{th}$ rule is:
\begin{equation} \label{eq:loss2}
\begin{aligned}
\mathcal{L}_i = & - \sum_{{\bf t}\in T_i^\prime}\log\left(\mathrm{sigmoid}\left(s_i\cdot \left(G_i\left({\bf t}\right)-\beta\right)\right)\right)\\
& - \sum_{{\bf t}\in T_i^{\prime\prime}}\log\left(1-\mathrm{sigmoid}\left(s_i\cdot \left(G_i\left({\bf t}\right)+\beta\right)\right)\right)
\end{aligned}
\end{equation}
where again $T_i^\prime$ and $T_i^{\prime\prime}$ are the two sets of training images that this rule distinguishes.
(\ref{eq:loss2}) is also the cross-entropy loss with poor man's adversarial training.

In the third step, parameters $b_i,1\leq i\leq 2632$ are learned jointly.
I again use poor man's adversarial training but, instead of a fixed $\beta$, I apply an image-dependent amount of adversarial adjustment on $G_i\left(\cdot\right)$'s.
Let $\beta_{\bf t}$ denote the adjustment amount for training image ${\bf t}$.
Let ${\bf o_\mathrm{ori}}\left({\bf t}\right)$ denote (\ref{eq:classi}) without adjustment and let ${\bf o_\mathrm{adv}}\left({\bf t}\right)$ denote that with adjustments of $\beta_{\bf t}$.
The $\beta_{\bf t}$ value is chosen such that ${\bf o_\mathrm{adv}}\left({\bf t}\right)$ barely classifies correctly, and it is periodically updated during the training process.
The loss function of the third step is:
\begin{equation} \label{eq:loss3}
\resizebox{.89\linewidth}{!}{$
\begin{aligned}
\mathcal{L} = & \mathrm{avg}_{\bf t}\left(\textrm{softmax-cross-entropy}\left({\bf o_\mathrm{adv}}\left({\bf t}\right),\textrm{label}_{\bf t}\right)\right) \\
& + \omega\cdot\mathrm{avg}_{\bf t}\left(\textrm{softmax-cross-entropy}\left({\bf o_\mathrm{ori}}\left({\bf t}\right),\textrm{label}_{\bf t}\right)\right)
\end{aligned}
$}
\end{equation}
where $\omega$ is a hyperparameter.
Intuitively, the image-dependent adversarial adjustments cause a uniform push to classify all images more robustly.
It's worth noting that, applying (\ref{eq:gen}), it can be shown that the softmax cross entropy of ${\bf o}\left({\bf t}\right)$ is equal to the log likelihood of generating the label of ${\bf t}$ if assuming a uniform prior-knowledge distribution.

\subsection{Results}

The pre-trained NBR MNIST 4-9 classifier is available at\\
\url{http://researcher.watson.ibm.com/group/10228}

Table~\ref{tbl:robust} compares the NBR MNIST classifier against those in \cite{mit,kolter,l2nnn}, which are publicly available.
I simply use their two logits for 4 and 9 to form binary classifiers.
Among them, the L2NNN classifier from \cite{l2nnn} is the state of the art in robustness as measured by $L_2$ metric.

To choose a meaningful $L_2$ $\varepsilon$, I measure $d\left({\bf t}\right)$, which is the distance from ${\bf t}$ to the nearest training image with a different label, and the \emph{oracle robustness} for ${\bf t}$ is $L_2$ radius $d\left({\bf t}\right)/2$.
Over the MNIST training set, the oracle robustness radius is above 3 for 51\% of images, above 2.5 for 79\%, and above 2 for 96\%.
Consequently $\varepsilon=2$ is the meaningful $L_2$ threshold when quantifying robustness of MNIST classifiers.

Robustness is measured by running four attacks: projected gradient descent (PGD) \cite{mit}, boundary attack (BA) \cite{boundary}, Carlini \& Wagner (CW) attack \cite{carlini} and seeded CW (SCW).
Foolbox \cite{foolbox} is used for PGD and BA; CW is original code from \cite{carlini}; SCW is a CW search with a starting point that is provided by a transfer attack, and is a straightforward variation of the CW code.
Iteration limit is 100 for PGD, 50K for BA, and 10K for CW and SCW.
A classifier is considered robust on an image if it remains correct under all four attacks.
Table~\ref{tbl:robust} shows that the NBR classifier has the best robustness, and also the best natural accuracy among all but the non-robust vanilla model.

\begin{table}[t]
\centering
\resizebox{.99\columnwidth}{!}{
\begin{tabular}{lcccccc}
\toprule
                & natural & PGD    & BA     & CW     & SCW    & robust \\
\midrule
Vanilla          & 99.7\% & 48.8\% &    0\% &    0\% &    0\% &    0\% \\
\citeauthor{mit} & 98.6\% & 97.8\% & 10.6\% & 47.8\% & 17.6\% &  1.3\% \\
Wong\&Kolter     & 98.9\% & 97.7\% & 15.9\% & 60.4\% & 28.3\% & 12.0\% \\
L2NNN            & 99.1\% & 94.4\% & 42.7\% & 41.3\% & 41.3\% & 41.3\% \\
NBR              & 99.1\% & 92.5\% & 57.2\% & 55.5\% & 55.3\% & 55.3\% \\
\bottomrule
\end{tabular}
}
\caption{Accuracies on natural and adversarial test images of 4 and 9 where the $L_2$-norm limit of distortion is 2.
Accuracies in the last column are under the best of four attacks for each image.}
\label{tbl:robust}
\end{table}

\begin{table}[t]
\centering
\resizebox{.7\columnwidth}{!}{
\begin{tabular}{lcc}
\toprule
        & natural & robust \\
\midrule
NBR                       & 99.1\% & 55.3\% \\
Markov random field \#1   & 97.6\% & 52.0\% \\
Markov random field \#2   & 90.0\% & 47.8\% \\
Gaussian naive Bayes \#1  & 98.6\% & 53.9\% \\
Gaussian naive Bayes \#2  & 69.4\% & 33.7\% \\
\bottomrule
\end{tabular}
}
\caption{Accuracies of ablation-study models.}
\label{tbl:ablation}
\end{table}

Table~\ref{tbl:ablation} presents empirical comparisons between NBR and other ensemble methods, by replacing belief-function arithmetic with Markov random fields (MRF) and Gaussian naive Bayes (GNB).
The difference between MRF \#1 and MRF \#2 is that the former uses $-\mathrm{sigmoid}\left(s_i\cdot G_i\left(\cdot\right)\right)$ as energy functions while the latter uses $-\log\left(\mathrm{sigmoid}\left(s_i\cdot G_i\left(\cdot\right)\right)\right)$; parameters $s_i$ are re-trained together with MRF's weight parameters; the same poor man's adversarial training of (\ref{eq:loss3}) is applied in training MRF models.
The difference between GNB \#1 and GNB \#2 is that the former uses $\mathrm{sigmoid}\left(s_i\cdot G_i\left(\cdot\right)\right)$ as features while the latter uses $G_i\left(\cdot\right)$.

\section{Related Work} \label{sec:liter}

As a formalism for reasoning with uncertainty, belief functions \cite{ds} have two distinct advantages: explicit modeling of the lack of knowledge and an elegant mechanism of combining multiple sources of information. 
Reasoning frameworks based on belief functions have been proposed \cite{gordon1985,baldwin1986,lowrance1986,laskey1988,dam1988,blp}, and they have varying degrees of similarity to the restricted framework of Section~\ref{sec:proto}.
There are another set of works that combine belief functions and fuzzy sets \cite{zadeh1979fuzzy,yen1990generalizing,denoeux2000modeling}, and the commonality between them and NBR is that my equation (\ref{eq:fuzzybel}), the mapping from a basic probability assignment to a belief function, coincides with that in \cite{zadeh1979fuzzy}.

These early works share some common weaknesses: where do rules come from, and where do uncertainty quantifications on the rules come from.
Relying on manual inputs is clearly not scalable.
To be fair, mainstream methods based on Bayesian networks \cite{pearlbook} or Markov random fields \cite{mrfbook} often have the same weaknesses.
Some have addressed the second weakness: for example, Markov logic networks \cite{mln} learn the weights on clauses.
Attempts to address the first weakness, e.g., by inductive logic programming \cite{muggleton1994,pilp}, are often limited.

Progresses in addressing the first weakness via automatic discovery of rules have emerged from the machine learning field.
In \cite{rbm,dbm}, Markov random fields, which can be viewed as compositions of non-symbolic rules, are learned from data.
In \cite{franca2014,dilp}, symbolic logic programs are learned from examples.
Another example is \cite{serafini2016} which defines a formalism of real-valued logic and thereby enables learning non-symbolic rules.
The training algorithm of NBR represents a new approach on this front.

Adversarial robustness is a well-known difficult problem \cite{szegedy,goodfellow,carlini}, and many remedies have been tried and failed \cite{carlini2017,athalye2018}.
For MNIST, if distortion is measured by the $L_\infty$ distance, there are a number of approaches \cite{kolter,raghu,towards} and in particular \cite{mit} achieves good $L_\infty$ robustness by adversarial training.
For $L_2$ robustness which is less understood and perhaps more difficult, before this work the state of the art is an L2NNN from \cite{l2nnn} with adversarial training.
It's worth noting that adversarial training alone does not work well for $L_2$ robustness \cite{towards,l2nnn,tradeoff}.
My hypothesis is that reasoning is a missing piece in previous works, and the NBR classifier is a demonstration that it's possible to break a classification task into simpler and smaller tasks, each of which is robustly solvable by an L2NNN, and that NBR can reason about the resulting many sources of evidence to reach a robust conclusion.

\section{Conclusions and Future Work}

This paper presents neural belief reasoner, which is a new generative model and a new approach to combine learning and reasoning.
Its properties are studied through two tasks: an unsupervised-learning task of reasoning with uncertainty, and a supervised-learning task of robust classification.
In the latter task, the MNIST classifier for 4 and 9 sets a new state of the art in $L_2$ robustness while maintaining over 99\% nominal accuracy.

An important future direction is improving the scalability of unsupervised learning.
For example, the first task uses an NBR with $F\left(\cdot\right)$ being identity function, and the complexity of exact calculus in unsupervised learning is exponential with respect to the number of rules.
To unlock its full potential, innovations would be needed for efficient inference and training algorithms, including but not limited to Monte Carlo methods, as well as efficient constraint-programming solvers.
There are also open questions from the application perspective, e.g., how to take advantage of NBR's sample-generation capability and how to leverage NBR for interpretability.

\bibliographystyle{named}
\bibliography{nbr}

\end{document}